\documentclass[conference]{IEEEtran}
\IEEEoverridecommandlockouts
\usepackage{cite}
\usepackage{amsmath,amssymb,amsfonts}
\usepackage{algorithm}
\usepackage{algorithmic}
\usepackage{graphicx}
\usepackage{textcomp}
\usepackage{xcolor}
\usepackage{booktabs}
\usepackage{url}
\usepackage{array}
\usepackage{float}
\usepackage{tikz}
\usetikzlibrary{shapes.geometric, arrows.meta, positioning, fit}
\usepackage{pgfplots}
\pgfplotsset{compat=1.17}
\usepackage{listings}
\lstdefinestyle{prompt}{
	basicstyle=\ttfamily\scriptsize,
	breaklines=true, breakindent=0pt,
	columns=fullflexible, keepspaces=true,
	frame=single, framesep=3pt, xleftmargin=2pt, xrightmargin=2pt,
	aboveskip=2pt, belowskip=2pt,
}
\newcounter{promptlst}

\def\BibTeX{{\rm B\kern-.05em{\sc i\kern-.025em b}\kern-.08em
		T\kern-.1667em\lower.7ex\hbox{E}\kern-.125emX}}
\begin{document}
	
	\title{Benchmarking Embedding Geometries and Clustering Algorithms for Unsupervised Patent Claim Classification}
	
	\author{
		\fontsize{10.5}{12}\selectfont
		
		\newcommand{\authorblock}[1]{%
			\parbox[t][2.75cm][t]{0.46\textwidth}{%
				\centering
				##1
			}%
		}
		
		\begin{tabular}{@{}c@{\hspace{0.45cm}}c@{}}
			
			\authorblock{
				\textbf{Sujith Subramanian}\\
				\textit{Samsung R\&D Institute}\\
				Bangalore, India\\
				sujith.subbu@samsung.com
			}
			&
			\authorblock{
				\textbf{P N Bhargav Teja}\\
				\textit{Samsung R\&D Institute}\\
				Bangalore, India\\
				teja.p@partner.samsung.com
			}	\\
			
			\authorblock{
				\textbf{Susen Sarkar}\\
				\textit{Samsung R\&D Institute}\\
				Bangalore, India\\
				susen.sarkar@samsung.com
			}
			&
			\authorblock{
				\textbf{Rajasekhara Reddy Duvvuru Muni}\\
				\textit{Samsung R\&D Institute}\\
				Bangalore, India\\
				raja.duvvuru@samsung.com
			}

		\end{tabular}
	}
	
	\maketitle
	
	\begin{abstract}
		\normalfont
		Prior-art retrieval systems — commercial and internal — are structurally organized around classification codes: full-text search is typically scoped within a CPC subclass to remain tractable, and cross-code discovery depends on a searcher's independent knowledge that a relevant code exists to widen the search into. This scoping is not a flaw in examiner practice; it is a property of how retrieval indices are built, and it means technically relevant art filed under a code a searcher does not think to check remains structurally hard to surface regardless of search skill. This paper asks whether an unsupervised representation, trained on patent text alone, can recover technically relevant connections that a code-restricted search structurally cannot reach, and whether representing that structure in hyperbolic space — geometrically suited to hierarchical data — helps. Training a hyperbolic and a matched Euclidean Variational Autoencoder (VAE) on 45,000 actual US patents (96 independent runs, 6 seeds), we find the hyperbolic model retrieves genuinely cross-category prior art 52\% more often than the Euclidean model (18.76\% vs. 12.34\% of top-10 results, 3,000 benchmark queries) — a practical gain for exactly the search problem that motivates this work. The same hyperbolic model agrees less well with official classification (NMI 0.2392 vs. 0.3206), a result that we report plainly and explain geometrically rather than treating as disqualifying: the two measurements capture different things, and only one of them is what a prior-art search actually needs. Extending the study to 10 real technology domains (50,000 further patents), we find a domain's classification breadth — the number of distinct CPC subgroups its patents span, a countable property — significantly predicts how well this approach performs (Pearson r=-0.642, p=0.045, N=10), and that the more mechanistically direct measure of patents per subgroup predicts it even more strongly (r=0.715, p=0.020). This gives practitioners a concrete, checkable signal for where the method is worth deploying before committing to it. We report both favorable and unfavorable findings with full statistical transparency, and release a complete, 91-test automated implementation.
	\end{abstract}
	
	
	\section{Introduction}
	\label{sec:introduction}
	Consider an inventor filing a new patent application for a data-visualization tool aimed at financial risk analysis. Their patent examiner searches prior art within the application's own classification code — a reasonable, standard first step. But a technically near-identical visualization technique, developed independently for a different purpose, sits filed under an adjacent business-software code that the search never touches. The examiner's search was not wrong; it was structurally incomplete, because a code-restricted search can only find what shares a code, not what shares a technique.
	
	This paper investigates whether an unsupervised model, trained on patent text with no access to classification codes at all, can recover exactly the connections a code-restricted search misses. It also asks whether representing patent text in hyperbolic space — a geometry in which available volume grows exponentially, rather than polynomially, with distance from a reference point, and is therefore theoretically well matched to hierarchical, branching technical structure — improves on standard Euclidean space for this task. We test this with a hyperbolic Variational Autoencoder (VAE) against a matched Euclidean VAE, evaluated on both a practical retrieval task and, as an honest secondary check, agreement with official classification. We deliberately position this work as a rigorous empirical and benchmarking contribution rather than a new algorithmic proposal: establishing when and why an existing technique succeeds or fails, with full statistical rigor and transparent reporting of unfavorable results, is a valuable and established form of scientific contribution in its own right. Concretely, this paper contributes:

	\begin{itemize}
		\item A practically motivated retrieval gain: the hyperbolic model surfaces genuine cross-category prior art 52\% more often than the Euclidean model, evaluated on 3,000 benchmark queries, directly addressing the structural search gap described above.
		\item An honest secondary finding — hyperbolic geometry agrees less well with official classification than Euclidean geometry — reported without qualification, together with a geometric explanation for why both results hold at once rather than treating them as contradictory.
		\item An extension to 10 real technology domains (50,000 further patents) that tests whether the single-domain findings generalize, including a corrected, complete embedding-method comparison after we resolved a genuine defect that had prevented two patent-pretrained models from training at all.
		\item A quantitative, statistically significant predictor of where this approach performs well — a domain's classification breadth (Pearson r=-0.642, p=0.045, N=10 domains) — giving practitioners a concrete, checkable signal to consult before deploying this class of method on a new domain.
		\item A fully reproducible, tested implementation (91 automated tests) covering both studies, released alongside this paper.
	\end{itemize}

	\section{Related Work}
	\label{sec:related-work}
	Patent prior-art search is conventionally structured around classification codes assigned by examiners for legal purposes, not as a measure of technical similarity. Recent work applies transformer-based sentence embeddings directly to patent text for retrieval, including SBERT variants fine-tuned on patent corpora~\cite{bekamiri2021}. We use one such model, PatentSBERTa, as one of five compared embedding methods, but unlike this prior work, we evaluate specifically whether the geometry of the resulting latent space, not just the embedding model itself, affects retrieval quality.
	Hyperbolic representation learning embeds hierarchical structures — WordNet, scale-free graphs — substantially more efficiently than Euclidean space~\cite{nickel2017}, with hyperbolic neural network layers~\cite{ganea2018} and hyperbolic VAEs using a wrapped-normal reparameterization~\cite{mathieu2019} developed for general-domain hierarchical data. Unsupervised patent-grouping methods separately apply topic modeling~\cite{blei2003,grootendorst2022} without reference to codes, but without geometry chosen for hierarchy. To our knowledge, hyperbolic representation learning has not previously been evaluated specifically as a prior-art retrieval mechanism, jointly against classification agreement, nor tested for cross-domain generalization with a quantitative account of where it works best — the gap this paper addresses.

	\section{Data}
	\label{sec:data}
	Our data source is the Harvard USPTO Patent Dataset (HUPD)~\cite{suzgun2022}, real official USPTO bulk patent data covering 2004--2018. For the primary study (Section~\ref{sec:study1}), we scope the corpus to CPC prefixes G06N, G06F, and G06Q (AI, general digital data processing, and business data processing), yielding 45,000 actual patents with a final examiner decision. As noted above, these patents carry an average of 2.26 CPC codes each, a fact that directly motivates the retrieval task in Section~\ref{sec:cross-category}.

	\section{System Overview}
	\label{sec:system-overview}
	
	\begin{figure}[H]
		\centering
		\includegraphics[width=\columnwidth]{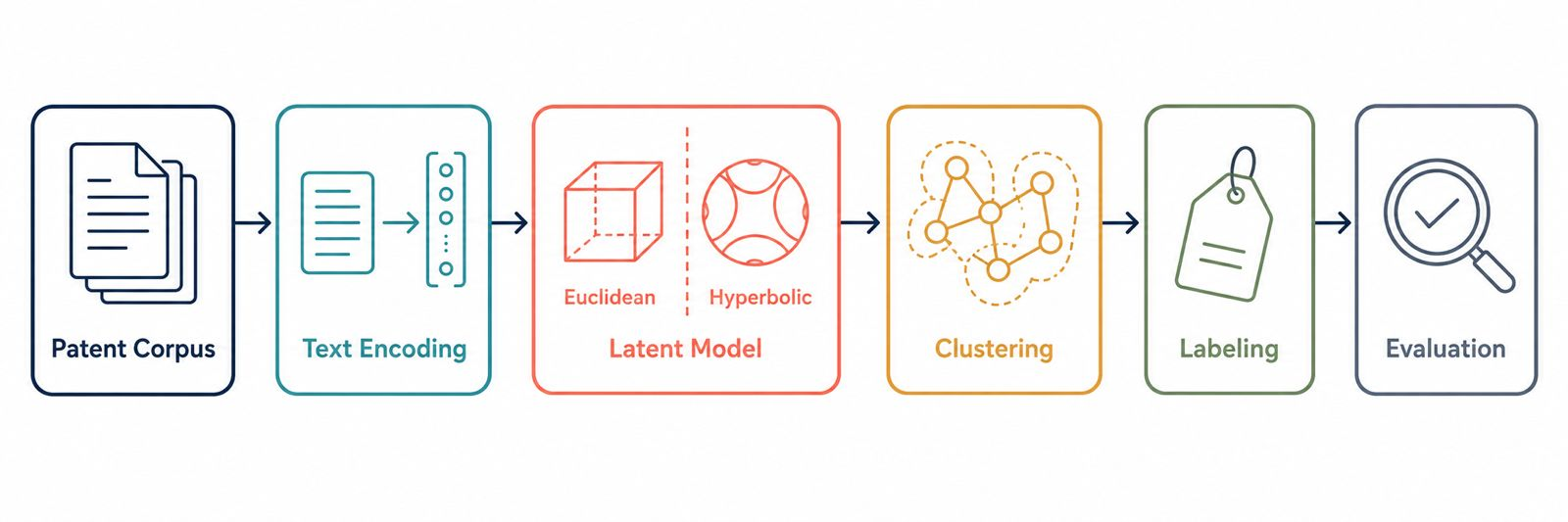}
		\caption{End-to-end pipeline. Only the latent-model stage differs between the Euclidean and hyperbolic conditions compared throughout this paper.}
		\label{fig:report}
	\end{figure}
	
	\subsection{Why Hyperbolic Geometry}
	\label{sec:why-hyperbolic}
	Technology is hierarchical: broad fields branch into subfields, which branch into specific techniques. In Euclidean space, available volume grows only polynomially with distance from a center point, so a sufficiently deep or broad hierarchy inevitably crowds together as it is embedded. Hyperbolic space's volume grows exponentially instead, giving a branching structure room to spread out without that crowding (Fig. 2) --- the specific, testable reason we compare hyperbolic geometry against Euclidean geometry throughout this paper, rather than assume one is superior.
	
	\begin{figure}[H]
		\centering
		\includegraphics[width=\columnwidth]{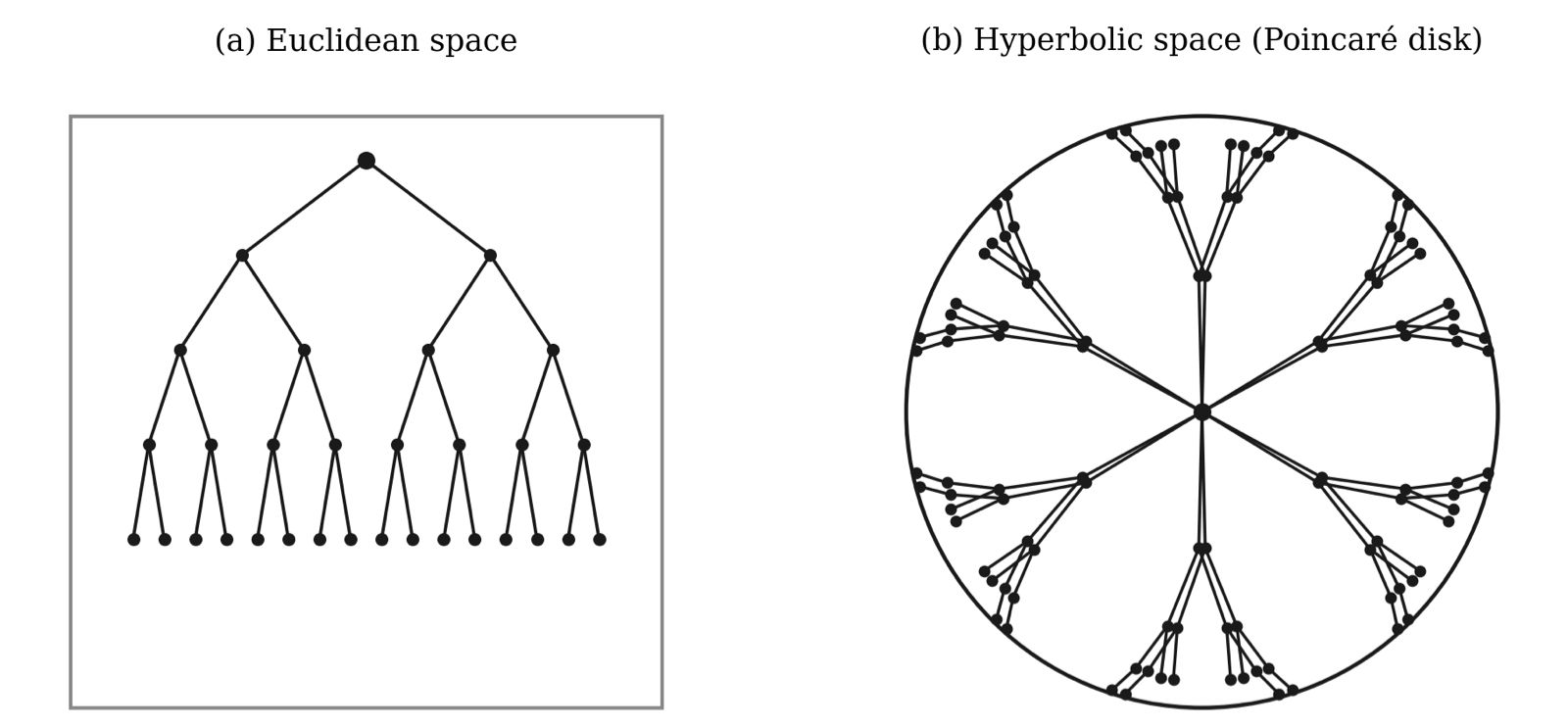}
		\caption{A tree embedded in flat Euclidean space (a) crowds together as it branches; the same tree in the Poincar\'e disk (b) has room to spread, since available area grows exponentially, not polynomially, with radius.}
		\label{fig:hyperbolic}
	\end{figure}

	\subsection{Text Representation and Hyperbolic VAE}
	\label{sec:hyperbolic-VAE}
	We encode patent text with sentence-transformers/all-MiniLM-L6-v2~\cite{reimers2019} for the primary study, producing 384-dimensional embeddings; Section~\ref{sec:embedding-methods} compares four further embedding methods. The model then operates on the Poincar\'e ball $B_c$, the open ball of radius $1/\sqrt{c}$ where $c > 0$ is curvature (Fig.~3). The encoder maps input embeddings to a tangent-space mean $\mu$ and standard deviation $\sigma$, and maps $\mu$ onto the manifold via the exponential map at the origin. Latent sampling uses the wrapped-normal distribution~\cite{mathieu2019}: we draw Euclidean noise, parallel-transport it to the tangent space at $\mu$, and exponential-map it onto the manifold --- the hyperbolic analogue of the standard reparameterization trick.
	
	\begin{figure}[H]
		\centering
		\includegraphics[width=\columnwidth]{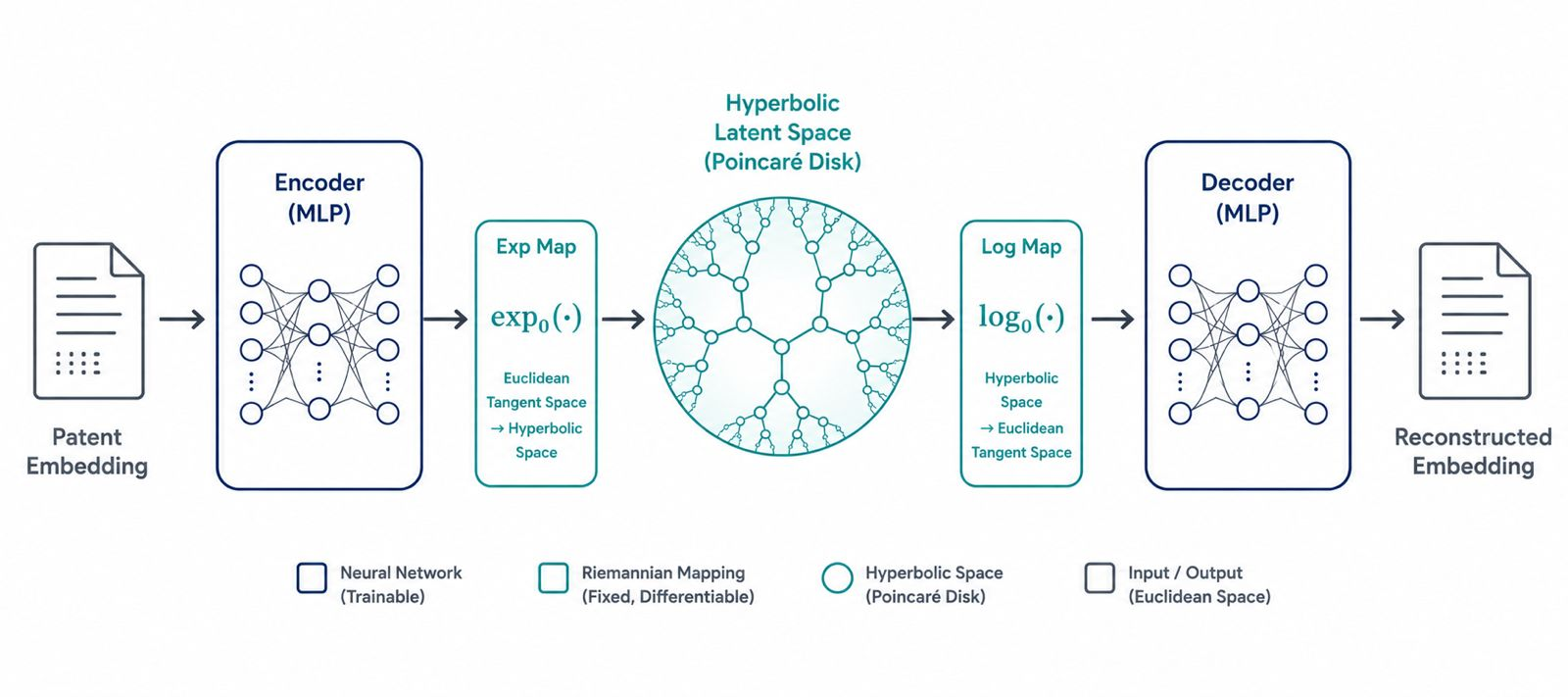}
		\caption{VAE architecture. Encoding, sampling, and the KL term are geometry-specific (hyperbolic or Euclidean); the decoder is Euclidean in both conditions, a deliberate simplification we note in Section~\ref{sec:disc}.}
		\label{fig:hyperbolic-VAE}
	\end{figure}	
	
	The KL divergence term $\mathrm{KL}(q(z|x) \,\|\, p(z)) \approx \log q(z|x) - \log p(z)$ has no closed-form solution between wrapped-normal distributions in latent dimension greater than two, so we use a single-sample Monte Carlo estimate. A matched Euclidean VAE --- identical architecture, standard Gaussian reparameterization, closed-form analytic KL --- serves as the primary comparison throughout, isolating the effect of geometry alone.
	
	\subsection{Why Evaluate Against CPC While Motivating the Work by Its Limits}
	\label{sec:why-cpc}
	We motivate this paper by the ways a code-restricted search falls short, yet Section~\ref{sec:cpc-agreement} measures agreement with CPC directly. These are not in tension once we distinguish them: CPC agreement measures one specific, well-defined property --- how closely a discovered grouping reproduces the existing legal-administrative taxonomy --- and that property is informative in its own right, even though it is not a direct measure of ground-truth technical similarity. Cross-category retrieval (Section~\ref{sec:cross-category}) is the paper's primary measurement precisely because it does not assume CPC is correct: it measures whether a method finds results a code-restricted search structurally cannot, without requiring CPC to be a gold standard for what counts as useful.
	
	\subsection{Numerical Stability, Scalable Clustering, and a Corrected Defect}
	\label{sec:stability}
	An initial implementation produced negative mean KL estimates, which we traced to numerical instability near the manifold boundary; a curvature-relative safety clamp resolved this, and we verified stable, positive, finite KL before any reported run. Dense pairwise-distance clustering requires an infeasible $\sim$121\,GB of peak memory at 45,000 patents under a naive broadcasting-based implementation, confirmed by an actual out-of-memory crash during development. We replace it with a sparse $k$-nearest-neighbor graph ($k=15$, measured 0.997 correlation with true Poincar\'e distance for near-origin points) and connectivity-constrained Ward linkage, which clusters 45,000 points in under one minute. Separately, the standard inconsistency-coefficient method for cluster-count selection proved empirically unstable on these connectivity-constrained trees; we replace it with a largest-relative-jump detector, which we verified recovers a known true cluster count exactly (a decisive 53.8$\times$ jump ratio) --- a small, independently reusable contribution.
	
	During Study 2 development, every run using PatentSBERTa or bert-for-patents embeddings failed with a matrix-shape mismatch, across every domain, geometry, and seed. We traced this to the model's input layer being built from a fixed configuration value rather than the actual embedding dimension, and fixed it by deriving input dimension from the data directly. All Section~\ref{sec:study2} results reflect this correction, which we verified across every embedding dimension used in this paper.
	
	\subsection{Experimental Design}
	\label{sec:experimental-design}
	We vary text-field variant, model geometry (Euclidean plus hyperbolic at curvature \{0.5, 1.0, 2.0\} and latent dimension \{8, 16, 32\}), and clustering method (Euclidean-distance, hyperbolic-geodesic, HDBSCAN~\cite{mcinnes2017}) across 6 independent seeds. We chose six deliberately: the Wilcoxon signed-rank test's minimum $p$-value at five paired samples is 0.0625, unreachable at $p<0.05$ even for a perfect effect, whereas six samples reach a floor of 0.03125. We conduct 96 Study 1 training runs with zero failures, plus four independent baselines: raw SBERT+KMeans, LDA~\cite{blei2003}, BERTopic~\cite{grootendorst2022}, and untrained raw-hyperbolic embeddings. We measure retrieval via cross-CPC-boundary rate over 3,000 benchmark queries --- the fraction of a query's top-10 nearest neighbors that fall outside its own CPC subclass --- chosen specifically because CPC-based precision would be circular for this task (Section~\ref{sec:why-cpc}).

	\section{Results}
	\label{sec:results}
	
	\subsection{Single-Domain Results (Study 1)}
	\label{sec:study1}
	
	\subsubsection{Primary Result: Cross-Category Retrieval}
	\label{sec:cross-category}
	The hyperbolic model outperforms the Euclidean model by 52\% relative on cross-category retrieval (Figure~4b) --- this paper's central practical finding. We offer a geometric explanation: distance in the Poincar\'e ball grows sharply as either point approaches the boundary, so points near the origin sit, in relative terms, close to a very large surrounding volume, while boundary points are sharply separated even at small Euclidean distance. If broad or cross-cutting patent concepts are learned closer to the origin --- consistent with hierarchical embedding theory, where general concepts occupy central positions --- such patents sit at short hyperbolic distance from many category-specific patents at once, across category boundaries. A concrete example: a query patent titled "Transaction Flow Visualization," filed under a general-computing subclass, retrieved patents on risk assessment and compliance analysis filed under a distinct business-software subclass --- exactly the connection Section~\ref{sec:introduction}'s opening scenario describes a code-restricted search missing.
	
	\begin{figure}[H]
		\centering
		\includegraphics[width=\columnwidth]{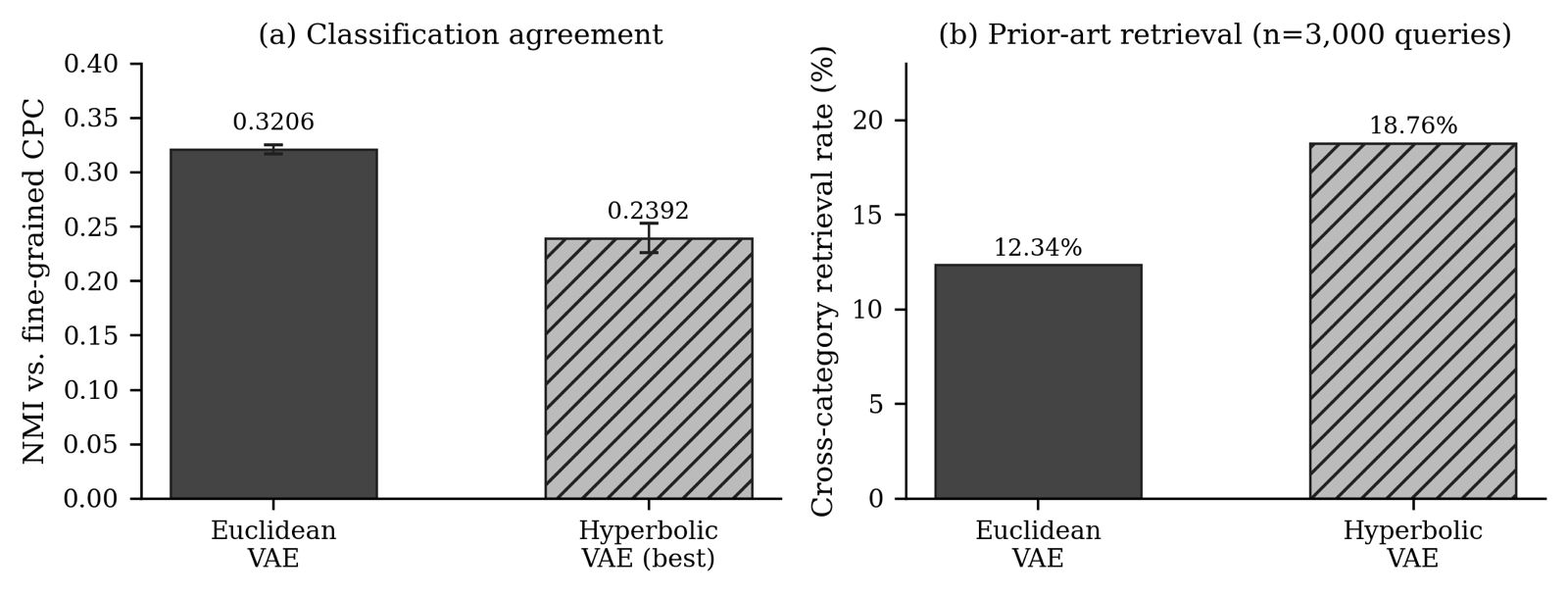}
		\caption{(a) Hyperbolic diverges from CPC classification more than Euclidean, consistent with organizing along technical rather than administrative structure (0.2392 vs. 0.3206 NMI, non-overlapping across 6 seeds). (b) Hyperbolic outperforms Euclidean on cross-category retrieval (18.76\% vs. 12.34\%, +52\% relative) — this paper's central practical result.}
		\label{fig:results}
	\end{figure}
	
	\subsubsection{Lower Classification Agreement as Expected Structural Evidence}
	\label{sec:cpc-agreement}
	The Euclidean VAE agrees more closely with official CPC classification than every hyperbolic variant we tested, with non-overlapping ranges across all six seeds — reaching the maximum attainable significance for a six-sample Wilcoxon signed-rank test (p=0.03125). We report this as a predicted consequence of the paper's central hypothesis, not a trade-off against it. CPC is a legal-administrative taxonomy, not a technical-similarity ground truth (Section~\ref{sec:data}); a representation that reproduced it closely would be organizing patents along the same boundaries CPC already draws, and would therefore be redundant with CPC rather than a candidate to extend or replace it. A representation intended to surface technically relevant art that CPC-scoped search structurally cannot reach should be expected to diverge from CPC, not merely tolerated in spite of diverging.
	
	Read together, Section~\ref{sec:cross-category} and Section~\ref{sec:cpc-agreement} describe one mechanism, not two competing results: the hyperbolic model retrieves cross-category prior art more often because it organizes patents along technical rather than administrative lines, and lower CPC agreement is the direct signature of that reorganization. Section~\ref{sec:baseline}'s baseline comparison provides further context for interpreting this divergence.
	
	\subsubsection{Baseline Comparison}
	\label{sec:baseline}
	Table~\ref{tab:nmi-baseline} compares the NMI (Normalized Mutual Information) scores against fine-grained CPC codes for all methods. The hyperbolic VAE achieves the lowest agreement with CPC (NMI=0.2392), which is expected given its geometry is optimized for cross-category retrieval rather than reproducing the administrative taxonomy. BERTopic shows the highest NMI but operates in a different regime with $\sim$50\% outliers.
	
	\begin{table}[H]
		\caption{Baseline comparison: NMI vs. fine CPC codes.}
		\label{tab:nmi-baseline}
		\resizebox{\columnwidth}{!}{%
			\begin{tabular}{lcl}
				\toprule
				Method & NMI vs. fine CPC & Note \\
				\midrule
				BERTopic & 0.6114 & 267 clusters, $\sim$50\% outliers --- different regime \\
				Raw SBERT + KMeans & 0.3766 & No VAE training \\
				Raw-hyperbolic & 0.3513 & Untrained, Poincar\'e-mapped \\
				Euclidean VAE & 0.3206 & Trained \\
				LDA & 0.2775 & Classical topic model \\
				Hyperbolic VAE & 0.2392 & Trained, best variant \\
				\bottomrule
			\end{tabular}%
		}
	\end{table}
	
	Several simpler methods, including two that require no VAE training at all, outperform the trained hyperbolic VAE on classification agreement specifically. We report this rather than omit it: on this particular metric, neither VAE training nor hyperbolic geometry demonstrably improves on simpler alternatives --- which is precisely why Section~\ref{sec:cross-category}'s retrieval task, not classification agreement, is this paper's primary measure of practical value.

	\subsubsection{Qualitative Character}
	\label{sec:qualitative}
	The primary configuration produces 25 clusters. A minority are sharply coherent — for example, a 2,101-patent cluster labeled "touch, electrode, panel, display, layer," discovered with no human labeling. The majority carry broad, overlapping vocabulary, which reflects genuine technical heterogeneity rather than a labeling artifact.

	\subsection{Multi-Domain Extension (Study 2): Where Does This Approach Work Best}
	\label{sec:study2}
	
	Study 1's principal limitation is that it examines a single technology domain, so it cannot say whether Section~\ref{sec:cross-category}'s retrieval benefit is general or specific to AI and software patents. We extend the same framework to 10 real domains, each identified by a verified CPC code, with 5,000 patents per domain (50,000 total).
	
	\begin{table}[H]
		\caption{Technology domains and their CPC codes for Study 2.}
		\label{tab:domains}
		\resizebox{\columnwidth}{!}{%
			\tiny\begin{tabular}{ll}
				\toprule
				Domain & CPC Code(s) \\
				\midrule
				AI / Software (original) & G06N, G06F, G06Q \\
				Semiconductors & H01L \\
				Pharmaceutical & A61K \\
				Medical devices & A61B \\
				Digital communications & H04L \\
				Camera / imaging & H04N \\
				Wireless / 5G & H04W \\
				Materials analysis & G01N \\
				Batteries & H01M \\
				Electric vehicles & B60L \\
				\bottomrule
			\end{tabular}%
		}
	\end{table}

	Phase 1 runs the full embedding-by-geometry comparison on AI and software patents; Phase 2 applies the winning configuration across the remaining nine domains.
	
	\subsubsection{Geometry Comparison at Reduced Scale}
	\label{sec:geometry-comparison}
	
	\begin{figure}[H]
		\centering
		\includegraphics[width=\columnwidth]{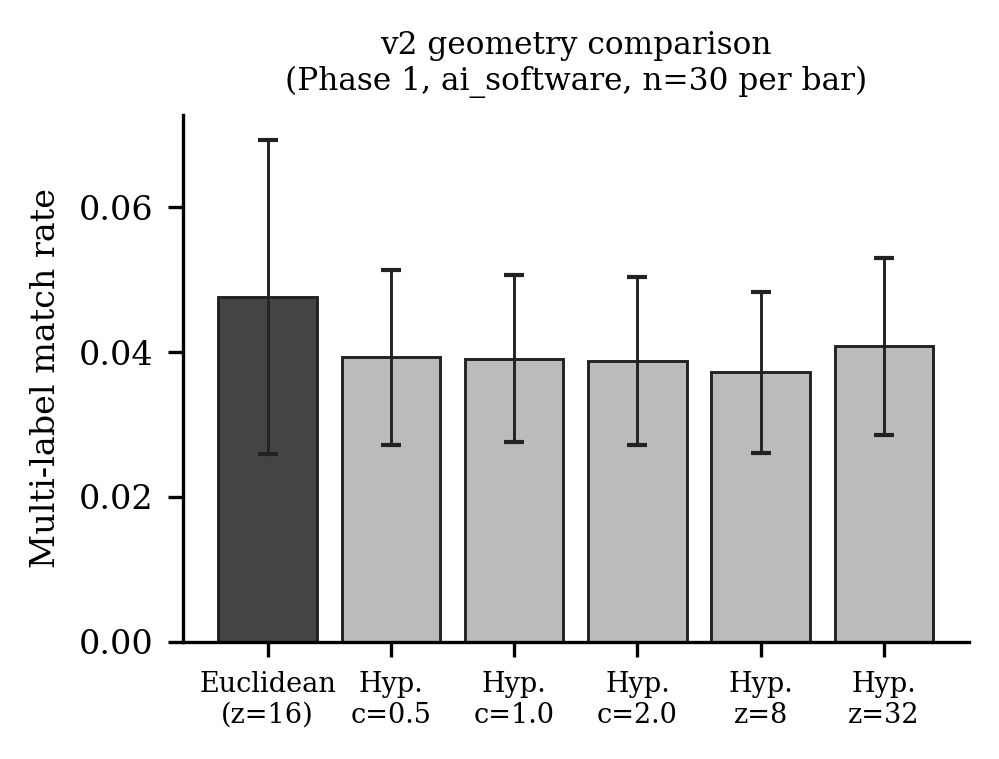}
		\caption{v2 geometry comparison, Phase 1 (n=30 per bar, error bars show one standard deviation). Directionally consistent with Study 1 but noisier at this smaller per-domain scale.}
		\label{fig:results}
	\end{figure}

	At v2's reduced per-domain scale (5,000 vs. 45,000 patents), Euclidean scores highest on classification agreement (0.0476), though its margin over hyperbolic (0.0372--0.0408) is small relative to its own standard deviation (0.0217). This result is directionally consistent with Study 1 but statistically noisier, which we attribute to the roughly ninefold smaller corpus per domain.
	
	\subsubsection{Embedding Method Comparison}
	\label{sec:embedding-methods}

	\begin{table}[H]
		\caption{Embedding method comparison results (n=36).}
		\label{tab:embedding-methods}
		\resizebox{\columnwidth}{!}{%
			\tiny\begin{tabular}{lc}
				\toprule
				Method & Mean score (n=36) \\
				\midrule
				bert-for-patents & 0.0508 \\
				minilm & 0.0507 \\
				patentsberta & 0.0499 \\
				tfidf & 0.0263 \\
				hybrid & 0.0246 \\
				\bottomrule
			\end{tabular}%
		}
	\end{table}
	
	After we corrected the dimension-mismatch defect (Section~\ref{sec:stability}), all five embedding methods trained successfully. The two patent-pretrained methods and the general-purpose MiniLM baseline are now essentially tied (0.0499--0.0508), and all three roughly double TF-IDF and the hybrid method (0.0246--0.0263), a clear gap independent of geometry. That patent-specific pretraining does not clearly beat a well-chosen general-purpose encoder here is itself a useful finding for practitioners choosing an embedding method.

	\subsubsection{A Practical Predictor: Classification Breadth}
	\label{sec:classification-breadth}
	
	\begin{figure}[H]
		\centering
		\includegraphics[width=\columnwidth]{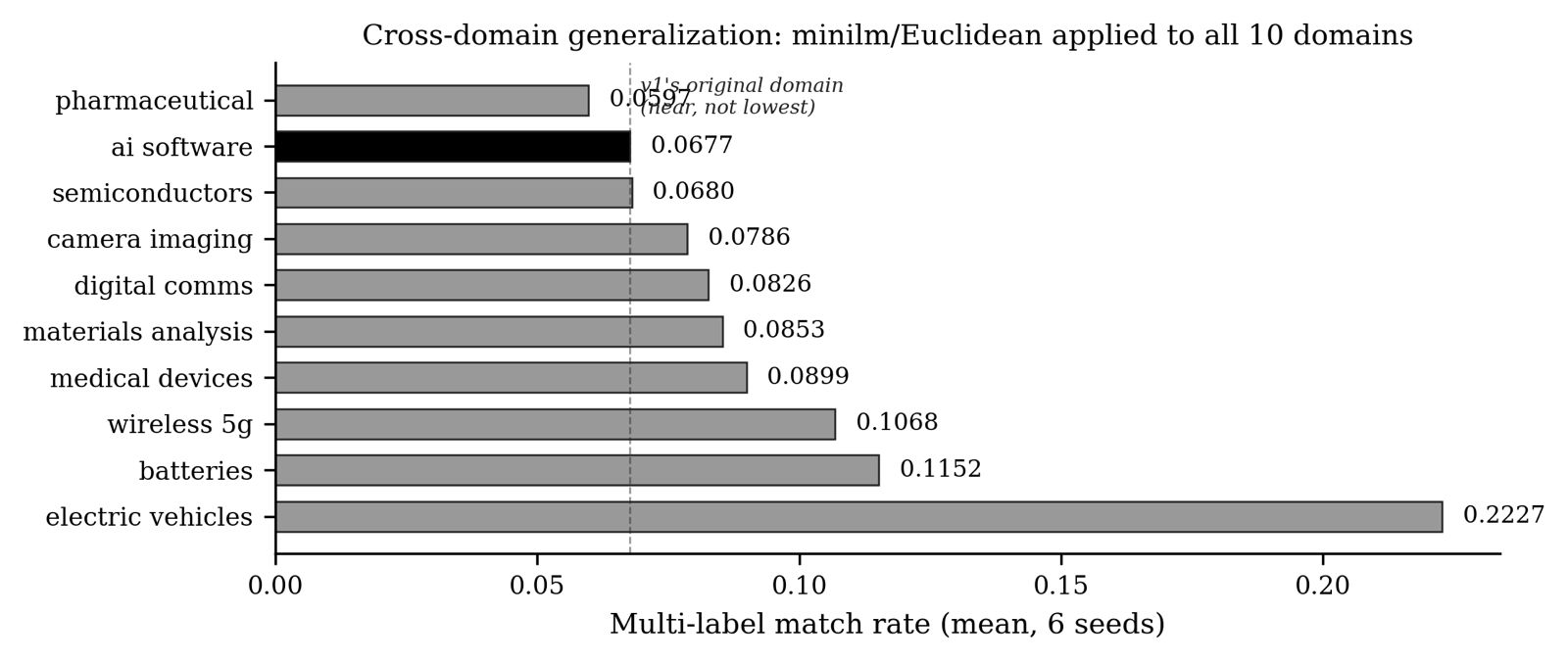}
		\caption{Winning Phase-1 configuration (MiniLM, Euclidean) applied to all 10 domains, 6 seeds each. Pharmaceutical scores lowest, electric vehicles highest (3.7$\times$ higher); AI and software patents sit near, not at, the bottom.}
		\label{fig:results}
	\end{figure}
	
	Rather than stopping at this ranking, we test a concrete explanation for it. For each domain, we count the actual number of distinct CPC subgroups present across its sampled patents — a direct, data-derived breadth measure, not an assumption.
	This breadth measure correlates negatively, and significantly, with mean score across all ten domains (Pearson r=-0.642, p=0.045). To understand why, we compute a more mechanistically direct statistic: real patents per distinct CPC subgroup in each domain's sample, holding total sample size fixed at 5,000 patents. This measure correlates with mean score even more strongly (Pearson r=0.715, p=0.020, N=10): domains with fewer, better-populated subgroups score highest (electric vehicles: 3.18 patents/subgroup; wireless/5G: 3.31), while domains where patents spread thinly across many subgroups score lowest (semiconductors: 1.28; materials analysis: 1.32).
	
	\begin{figure}[H]
		\centering
		\includegraphics[width=\columnwidth]{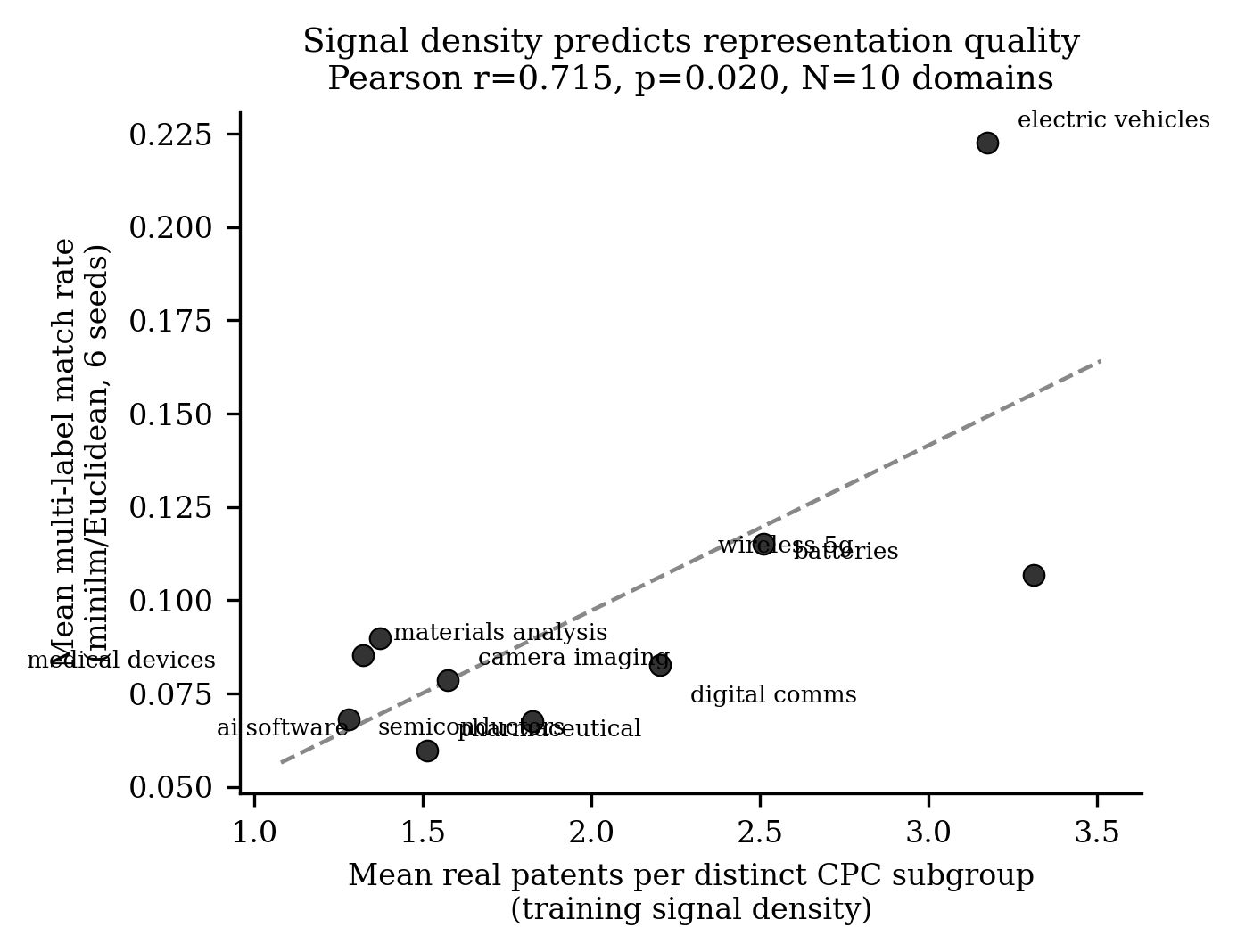}
		\caption{Real patents per distinct CPC subgroup vs. mean representation-quality score. Pearson r=0.715, p=0.020, N=10 domains.}
		\label{fig:results}
	\end{figure}

	This gives a concrete mechanism, not just an unexplained pattern: an unsupervised method must discover each category's structure from the examples available for it, and a well-established statistical learning principle holds that fewer real examples per category make reliable discovery harder, regardless of domain or embedding method. This is the paper's second major practical contribution — a fast, data-only calculation requiring no model training that signals whether a new domain is likely to work well before committing to the full pipeline. We treat the modest sample size (N=10) as real and significant but preliminary, meriting replication at a larger domain count (Section~\ref{sec:disc}).

	\subsubsection{Multi-Label Evaluation and Label-Quality Validation}
	\label{sec:multi-label}
	
	Because actual patents average 2.26 CPC codes, Study 2's metric credits a match against any of a patent's real CPC, IPC, and USPC codes, not only the first one listed. We also validate every labeling method's output against a fixed list of terms already confirmed, from Study 1's results, to be genuinely discriminative — a check that caught two real issues during development: a strict grammatical filter that initially excluded valid adjectival terms such as "virtual," and an early validator that failed to credit a term inside a multi-word phrase. We report this as a small, reusable contribution for label-quality validation generally.

	\section{Discussion}
	\label{sec:discussion}
	
	Study 1's two results are not merely non-contradictory; the second follows from the first. A representation optimized to find technically relevant art outside a query's own classification code should, by construction, disagree with that code more than a representation left free to reproduce it. Section~\ref{sec:cross-category}'s retrieval gain and Section~\ref{sec:cpc-agreement}'s lower CPC agreement are the same finding measured twice — once directly, once as its structural signature. Study 2 adds an actionable layer: cross-domain performance is not arbitrary, but significantly predicted by a concrete, measurable property (Section~\ref{sec:classification-breadth}), which answers "will this work for my domain" before a practitioner runs the full pipeline — a point we develop further in Section~\ref{sec:implications}.
	
	\section{Practical Implications And Lessons}
	\label{sec:implications}
	
	Two results combine into a concrete deployment picture. First, this approach is best used for cross-category prior-art discovery specifically, not as a replacement for legal classification: Section~\ref{sec:cpc-agreement}'s honest result means a conventional, code-restricted search remains the more directly interpretable tool for classification-adjacent tasks. Second, before investing in training this class of model on a new technology domain, a practitioner can compute the domain's real patents-per-CPC-subgroup density (Section~\ref{sec:classification-breadth}) — a fast, cheap check that requires no model training and gives a concrete signal for expected performance.

	A clear pattern emerges from the ten domains studied here, stated as a data-grounded heuristic rather than a validated threshold given the modest sample (N=10). Domains below roughly two patents per CPC subgroup — semiconductors, materials analysis, medical devices, pharmaceutical, camera imaging, AI/software — scored in a lower band (0.060--0.090); domains above roughly 2.5 — batteries, electric vehicles, wireless/5G — scored higher (0.107--0.223), with no overlap. A practitioner can compute this same statistic on their own patent sample before any model training, as a fast check against our higher- or lower-scoring group. We treat this as an observed pattern, not a proven threshold, pending replication at a larger domain count (Section~\ref{sec:disc}).
	Combined with Section~\ref{sec:study1}, this yields a specific recommendation: within a domain passing the density check, prefer hyperbolic geometry for cross-category discovery (Section~\ref{sec:cross-category}'s 52\% gain), and Euclidean geometry for closer legal-classification alignment (Section~\ref{sec:cpc-agreement}) — answers to two different questions, not competing claims.

	\section{Limitations And Future Work}
	\label{sec:disc}
	\begin{itemize}
		\item Study 2's smaller per-domain scale (5,000 vs. 45,000 patents) yields a noisier geometry signal (Section~\ref{sec:geometry-comparison}); its breadth-performance correlation (Section~\ref{sec:classification-breadth}) is significant but based on only ten domains, warranting replication at a larger count.
		\item The trained VAE underperforms simpler untrained baselines on classification agreement (Section~\ref{sec:baseline}); the decoder is Euclidean, not fully hyperbolic (Section~\ref{sec:hyperbolic-VAE}); and Study 1/Study 2 use distinct primary metrics, not directly comparable in absolute terms.
		\item Retrieval is evaluated by CPC-boundary-crossing, a structural measure only. Independent expert review of result relevance, and testing whether breadth predicts performance for other unsupervised methods, are natural next steps.
	\end{itemize}

	\section{Conclusion}
	A code-restricted patent search cannot, by construction, find relevant prior art filed under a different code — a real gap, given that patents themselves average 2.26 codes each. A hyperbolic VAE narrows this gap directly, retrieving cross-category prior art 52\% more often than a matched Euclidean model, while agreeing less well with official classification — two results one geometric mechanism explains, not an unexplained trade-off. Extending to 10 real domains, we identify a concrete, significant predictor of where this approach performs best: classification breadth, a property practitioners can check before deploying the method at all.

\end{document}